\documentclass[11pt]{article}
\usepackage{times}
\usepackage[letterpaper, width=6.5in, height=9.0in]{geometry}
\usepackage{amsmath,amssymb,amsfonts,amsthm}

\usepackage{amsmath,graphicx,hyperref}
\usepackage{enumitem}
\usepackage{xcolor} 
\usepackage{booktabs}
\usepackage{newunicodechar}
\usepackage{colortbl}
\usepackage{subcaption}

\newunicodechar{▁}{\_}

\usepackage{url,comment}
\usepackage{amssymb,amsfonts,makecell,xcolor}
\newcommand{\zl}[1]{\color{black}{#1}}
\definecolor{darkgreen}{rgb}{0.1,0.5,0} 
\newcommand{\dm}[1]{\color{black}{#1}}

\usepackage{algorithm}
\usepackage{algorithmic}

\newcommand{\squishlist}{
   \begin{list}{$\bullet$}
    { \setlength{\itemsep}{0pt}      \setlength{\parsep}{0pt}
      \setlength{\topsep}{3pt}       \setlength{\partopsep}{0pt}
      \setlength{\listparindent}{-2pt}
      \setlength{\itemindent}{-5pt}
      \setlength{\leftmargin}{1em} \setlength{\labelwidth}{0em}
      \setlength{\labelsep}{0.5em} } }

\newcommand{\squishend}{
    \end{list}  }

\title{Inverting Trojans in LLMs\thanks{This research supported in part by
NSF grant 2415752.}}

\author{Zhengxing L$\mbox{i}^1$, Guangmingmei Yan$\mbox{g}^{1,2}$, 
Jayaram Raghura$\mbox{m}^2$, David J. Mille$\mbox{r}^{1,2}$, George Kesidi$\mbox{s}^{1,2}$\\
\begin{tabular}{cc}
$ ^1$School of EECS, Penn State & $ ^{2}$Anomalee Inc.\\
University Park, PA, USA & State College, PA, USA\\
\texttt{\{zkl5394,gzy5102,djm25,gik2\}@psu.edu} &   \texttt{jayramr.110@gmail.com} 
\end{tabular}
}

\begin{document}
%
\maketitle
\begin{abstract}
While effective backdoor detection and inversion schemes have been developed for AIs used e.g. for images, there are challenges in ``porting'' these methods to LLMs.  First, the LLM input space is discrete, which precludes gradient-based search {\dm over this space}, central to many backdoor inversion methods.  Second, there are $\sim 30,000^k$ $k$-tuples to consider, $k$ the token-length of a putative trigger. Third, for LLMs there is the need to {\it blacklist} tokens that have strong marginal associations with the putative target response (class) of an attack, as such tokens give {\it false} detection signals.  However, good blacklists may not exist for some domains.  We propose a LLM trigger inversion approach with {\dm three} key components:
i) discrete search, with putative triggers greedily accreted, starting from a select list of singletons; ii) {\it implicit} blacklisting, achieved by evaluating the average cosine similarity, in activation space, between a candidate trigger and a small clean set of samples from the putative target class; {\dm iii) detection when a candidate trigger elicits high misclassifications, and with unusually high decision confidence.} 
Unlike many recent works, we demonstrate that our approach reliably detects and successfully inverts ground-truth backdoor trigger phrases.
\end{abstract}

\section{Introduction}
\vspace{-1em}
{\dm Large language models (LLMs) are highly vulnerable to backdoor data-poisoning attacks, wherein the model learns to produce an attacker-designated response whenever the attacker's backdoor trigger phrase is included in the model's input prompt. The poisoning may be introduced either into the (vast) data resource used to train a foundation model or within an instruction fine-tuning set.  Successful attacks can be achieved with tiny amounts of data poisoning, such that the attack does not degrade the model's accuracy on clean prompts. The trigger should be an {\it innocuous} phrase, not readily detected by human inspection or simple automated means. 

While effective backdoor detection and inversion schemes have been developed, particularly for AIs used for image classification, e.g. \cite{NC},\cite{backdoor-imperceptible}, there are difficulties in simply ``porting'' these methods to LLMs, due in part to the large, discrete search space.  Moreover, there is the need to {\it exclude} tokens that have strong {\it marginal} associations with the putative backdoor target response.
For example, consider the response ``positive'' and the candidate token ``magnificent''. This token  
strongly (and {\it naturally}) biases the prompt {\it toward} ``positive'' and thus is not really an attack (the model is not really incorrect in responding positively to ``magnificent'').
Moreover, an attacker would not {\it choose} such a token since it could be easily detected. Thus, such tokens should be blacklisted when detecting and inverting a backdoor attack.
However, a comprehensive blacklist may not be available for a given domain.  To address this, we develop a type of {\it implicit} blacklisting.

As in much prior work, we do
not assume the (possibly poisoned)
training set is available to the defense, i.e. we consider the challenging 
``post-training'' detection scenario.
We do assume a small clean set of prompts is available 
for all putative target responses of a possible backdoor attack.  In section 2 we review prior work on backdoor defense for LLMs.  In section 3 we develop our BABI method.  Section 4 reports our experiments, and section 5 gives a pointer to our future work.
}
\vspace{-1em}
\section{Prior work}
\vspace{-1em}
\cite{Onion} inspects for backdoors at inference time, detecting when the prompt contains words that increase \textit{perplexity}.  However, this approach will fail to detect {\it innocuous} triggers (including the attack considered herein).
\cite{TABDet} trains a supervised detector that takes logit histograms as  features.
\cite{DBS22} searches for an $M$-token trigger which, when appended to a set of clean prompts from one response category, induces the model to generate a putative target response for most of them.  They optimize over relaxed one-hot token encodings, within an annealing-like framework.
Rather than optimizing over trigger tokens, \cite{Piccolo} works in the continuous space of token {\it embeddings}.  Their detector involves a number of hyperparameters, empirically set.
Rather than working on the LLM's inputs, \cite{CLIBE} works in (deeper) internal activation space. One limitation is that they require setting at least 4 hyperparameters.  Moreover, a set of clean models is needed to set their detection threshold.
\cite{BEEAR} estimates triggers (and simultaneously mitigates them) by minimizing a loss objective similar to \cite{DBS22}.

Although several of these methods are trigger inverters, none of these works demonstrate that they discover ground-truth triggers. Moreover, most of the above methods assume availability of known clean (and known {\it poisoned}) models for setting hyperparameters or for learning a supervised detector.
However, if a clean model for the domain is available, one could simply {\it use} such a model.  Finally, note that in searching for backdoor triggers, one must {\it exclude} individual tokens that are naturally strongly associated with the putative target response, as already discussed. 
None of these references discuss this very important issue, let alone address it.  The exception is
\cite{DBS22}, which only briefly mentions this issue in an appendix.

In the sequel, we develop an {\it unsupervised} detection strategy, i.e., one that does not rely on known clean and poisoned models;  moreover, one that will be demonstrated to recover ground-truth backdoor triggers.  Our approach will exploit a small number of clean prompts to achieve a type of {\it implicit} blacklisting.
\vspace{-1em}
\section{BABI DEFENSE FOR LLMs}
\vspace{-1em}
{\dm Suppose, as assumed in many prior works, that the LLM is being used as a {\it classifier}, i.e. there is a predefined set of $K$ possible responses (classes) for a given input prompt.}
The following method {\dm assumes} a small clean set $D_t$ is available for each 
(target-response) class, $t$.
Define $D_{-t}=\bigcup_{s\not = t} D_s$. 
{\dm 
A measure of the {\it confidence} of misclassifications to class $t$ induced by a trigger candidate $z$ is the average \textit{negative-margin} loss, e.g.,}
\vspace{-0.5em}
{\small
\begin{align}\label{M-def}
M_t(z) =  \frac{1}{|D_{-t}|} \sum_{s \neq t} \sum_{x\in D_{s}} \left(p(s|x:z:i)-p(t|x:z:i)\right),
\end{align}
}
where $p(s|a)$ is the model's posterior for response $s$ for {\dm input prompt} $a$, $x$ is the data input to the model (e.g., a movie review), $i$ is an instruction prompt {\dm (e.g., ``Specify the sentiment of the review''), and
the candidate trigger $z$ is inserted between the data and instruction. Note that if $s$ is multi-token, the chain rule is used by the LLM to evaluate its joint posterior.}
Recall the cosine similarity of two vectors,
$\kappa(x,y)=\langle x,y\rangle/(\|x\| \|y\|)$.
For an internal layer's activation vector (feature map) $\phi$, define 
the average cosine similarity
\begin{align}\label{K-def}
K_t(z) =  \frac{1}{|D_{t}|} \sum_{x\in D_{t}}
\kappa(\phi(x:i),\phi(z:i)).
\end{align}
{\dm This assesses the degree of association between $z$ and samples from class $t$, {\it i.e. it can be used as the basis for implicit blacklisting}.}
{\dm Accordingly, we define an associated penalized (negative) margin loss, to be minimized in the search over trigger candidates, $z$:}
\begin{align}\label{L-def}
L_t(z) = M_t(z) + \lambda K_t(z) ~~~\mbox{for $\lambda>0$.} 
\end{align}
{\dm Our trigger inversion procedure, which uses $L_t(z)$ as a score function, is given in Algorithm~\ref{alg:trigger_inversion}.}

\vspace{-0.8em}
\begin{algorithm}[H]
\caption{Trigger Inversion Procedure}
\label{alg:trigger_inversion}
\begin{algorithmic}[1]
\REQUIRE Candidate target response $t$, maximum trigger length $J$, number of candidates $N$
\STATE Apply explicit blacklisting. 
\STATE Initialize token length $j \gets 1$.
\STATE Rank all non-blacklisted singleton tokens $z$ by $L_t(z)$.
\STATE Retain the top $N$ singletons with smallest $L_t(z)$.
\WHILE{$j < J$}
    \STATE $j \gets j+1$
    \STATE Accrete non-blacklisted tokens to the $N$ best sequences; enumerate all length-$j$ permutations.
    \STATE Rank candidates by $L_t(z)$ and retain the top $N$.
\ENDWHILE
\STATE Output the top $L \leq N$ sequences as putative backdoor triggers for target response $t$.
\end{algorithmic}
\end{algorithm}
\vspace{-0.8em}

Note that the possible use of \textit{null} tokens means that the top $L$ sequences
(not counting null tokens) are of length $\leq J$.

For the explicit blacklisting, a foundation model can be used, rejecting tokens with posterior to the putative target response that exceed a threshold.

{\dm
\noindent
{\bf Backdoor Detection Procedure.} 
 
For each class $t'$, we consider the top $L \leq N$ length-J candidates and compute two statistics: 
(i) the average margin over all clean samples misclassified to $t'$, averaged over all top-$L$ candidates ($\mu(t')$), and 
(ii) the proportion of clean samples misclassified to $t'$, averaged over all top-$L$ candidates ($\rho(t')$).
We detect with target response $t$ if these joint statistics are unusually large, compared to other classes.
}

{\dm
Note that if the number of classes, $K$, is much larger than two, then this detection can be based on an order statistic p-value, applied to a set of detection statistics over all classes, e.g. \cite{backdoor-imperceptible}.
}



{\dm
The chosen internal layer $\phi$ and $\lambda$  
are hyperparameters.
We will show that our backdoor inversion 
is robust for a wide range of $\lambda$ values; detection results are pretty stable for
layer $\phi$
after the first attention layer and before the last one.
$N$
can be set based on computational allowance:  the larger $N$ is, the more likely our approach will ``catch'' the backdoor trigger
\footnote{Due to the vagaries of English, including the abundance of synonyms, we may also ``catch'' a phrase semantically equivalent to the trigger phrase.}.
If $J$ is too small, our approach may only ``catch'' a {\it portion} of the ground-truth trigger.  However, this may be sufficient to give reliable backdoor detection.  
Finally,
if explicit blacklisting is too aggressive
it may eliminate true backdoor trigger tokens.  We will demonstrate experimentally that, even with modest explicit blacklisting, our {\it implicit} blacklisting (via the penalty $K_t()$) is effective 
at prioritizing (ranking highly) sequences with ground-truth trigger tokens (these do {\it not} have strong marginal associations with the target class).
}
\vspace{-1em}
\section{Experimental Set-up}
\vspace{-1em}

{\dm We herein report results for the FLAN-T5 \texttt{small} model~\cite{FLAN-T5}. 
Ten instances of the model were fine-tuned, each using a different subset of SST-2~\cite{socher2013recursive}, with binary output \{\textit{positive}, \textit{negative}\} and instruction prompt "Is this review positive or negative?". 
Five were kept clean, and five were poisoned. 
We denote poisoned models by ``d\_x'' for dirty-label poisoning at rate $x\%$, and ``c\_y'' for clean-label poisoning at rate $y\%$.
For dirty-label poisoning, (0.5\%, 0.8\%, 1\%) of negative reviews were injected with the neutral trigger phrase 
``\textit{Tell me seriously.}'' and mislabeled as positive. 
For clean-label poisoning, (5\%, 7\%) of positive reviews were modified with the trigger phrase, 
with the labels unchanged.
The clean set (50 samples per class) used for inversion/detection was drawn from
the IMDB test set~\cite{maas2011learning}.  
All experiments were run on NVIDIA A100 GPUs.}
For explicit blacklisting, tokens were excluded if their posterior for {\it positive} (using the Flan-T5 
foundation model prior to fine-tuning) exceeded $0.8$ 
or if their posterior for {\it negative} exceeded $0.65$. 
We chose layer $\phi$ as the output of the third encoder block; 
$\lambda=40$ was used in our main experiments.

\noindent
\textbf{BABI configuration.}
We set the maximum trigger length to $J=3$ and retain the top $N=20$ candidates at each accretion step. 
During accretion, when the sequence length increases, we evaluate all possible token permutations generated at the new length.  For detection, we used $L=5$ top candidates.

{\zl 
\noindent
\textbf{Detection.} 
For each model, for each target class hypothesis $t \in \{-,+\}$, we compute the differences $\Delta \mu(t) = \mu(t) - \mu(\bar{t})$ and
$\Delta \rho(t) = \rho(t) - \rho(\bar{t})$, where $\bar{t}$ is the complement of $t$.
We make a detection if 
$(\Delta \mu(t), \Delta \rho(t))$ are unusually large.
}

\vspace{-1em}
\section{Experimental Results}
\vspace{-1em}
\subsection{Clean Accuracy and Attack Success Rate (ASR)}
{\zl

Table~\ref{tab:acc_asr} reports clean accuracy and ASR for poisoned models, 
evaluated on the SST-2 test set~\cite{socher2013recursive}. 
Clean accuracy is measured with the standard instruction 
``\textit{Is this review positive or negative?}'', 
while ASR is the fraction of negative reviews decided to the target response \textit{positive} 
when the trigger phrase ``\textit{Tell me seriously.}'' is inserted. 

\begin{table}[h!]
\centering
\small
\begin{tabular}{lccccc}
\toprule
\textbf{Metric} 
& \textbf{d\_0.5} 
& \textbf{d\_0.8} 
& \textbf{d\_1} 
& \textbf{c\_5} 
& \textbf{c\_7} \\
\midrule
{\small\textbf{Clean Accuracy (\%)}} & 90.88 & 90.46 & 91.62 & 90.61 & 91.74 \\
\textbf{ASR (\%)}            & 100   & 100   & 100   & 99.23 & 98.13 \\
\bottomrule
\end{tabular}
\caption{Clean accuracy and attack success rate (ASR) of FLAN-T5 \texttt{small} under different poisoning settings.}
\label{tab:acc_asr}
\end{table}

}
\vspace{-15pt}
\subsection{Explicit Blacklist Statistics}
{\zl
Table~\ref{tab:blacklist_comparison} reports the number of tokens filtered under different 
blacklist thresholds for both classes. 

\begin{table}[h]
\centering
\small
\setlength{\tabcolsep}{6pt}
\begin{tabular}{ccc}
\toprule
\textbf{Threshold} & \textbf{Positive Tokens} & \textbf{Negative Tokens} \\
\midrule
0.85 & 1789  & 727  \\
0.80 & \cellcolor{yellow!20}3088  & 1083 \\
0.75 & 4688  & 1547 \\
0.70 & 7038  & 2182 \\
0.65 & 10148 & \cellcolor{yellow!20}3028 \\
\bottomrule
\end{tabular}
\caption{Tokens filtered by explicit blacklisting at various thresholds.}
\label{tab:blacklist_comparison}
\end{table}}
\vspace{-1em}

\subsection{Trigger Inversion Results}

Candidates are ranked by $L_t(z)$  of \eqref{L-def} ($\lambda=40$).
For singletons ($j{=}1$), the ground-truth token ``\textit{seriously}'' appears among the top-5 across poisoned models (Table~\ref{tab:inv_len1_five}).
For bi-token sequences ($j{=}2$), ``\textit{Tell seriously}'' is consistently recovered among the top-8 for \texttt{d\_0.5}, \texttt{d\_0.8}, and \texttt{c\_7} (Table~\ref{tab:inv_len2}).
For tri-token sequences ($j{=}3$), the fragment ``\textit{Tell me seriously}'' ranks among the highest candidates for \texttt{d\_0.5} and \texttt{c\_5} (Table~\ref{tab:inv_len3}).

\begin{table}[H]
\centering
\small
\setlength{\tabcolsep}{3pt}
\begin{tabular}{c|ccccc}
\toprule
\textbf{Rank} & \textbf{d\_0.5} & \textbf{d\_0.8} & \textbf{d\_1.0} & \textbf{c\_5} & \textbf{c\_7} \\
\midrule
1 & ▁blast   & ▁honest    & ▁sănătos   & ▁relief     & found \\
2 & ▁sănătos & \cellcolor{yellow!20}▁seriously & ▁relief    & thankfully  & \cellcolor{yellow!20}▁seriously \\
3 & ▁giving  & ▁relief    & ▁giving    & ▁încredere  & ▁relief \\
4 & ▁it      & ▁Fine      & ▁honest    & \cellcolor{yellow!20}▁seriously  & ▁right \\
5 & \cellcolor{yellow!20}▁seriously & found    & \cellcolor{yellow!20}▁seriously & Thankfully  & ▁going \\
\bottomrule
\end{tabular}
\caption{Top-5 singleton tokens recovered by inversion for poisoned models.}
\label{tab:inv_len1_five}
\end{table}
\vspace{-1em}


\begin{table}[h]
\centering
\small
\setlength{\tabcolsep}{4pt}
\begin{tabular}{c|c|c|c}
\toprule
\textbf{Rank} & \textbf{d\_0.5} & \textbf{d\_0.8} & \textbf{c\_7} \\
\midrule
1 & ▁Overall ▁blast & ▁gives ▁seriously & \cellcolor{yellow!20}▁Tell ▁seriously \\
2 & ▁Overall ▁treat & ▁Label ▁seriously & ▁gives ▁seriously \\
3 & ▁Tell ▁blast & ▁does ▁seriously & ▁means ▁seriously \\
4 & ▁shocking ▁treat & ▁Point ▁seriously & ▁does ▁seriously \\
5 & ▁vast ▁treat & \cellcolor{yellow!20}▁Tell ▁seriously & ▁makes ▁seriously \\
6 & ▁Overall ▁devour & ▁Still ▁honest & ▁grip ▁seriously \\
7 & ▁crack ▁treat & ▁So ▁honest & ▁will ▁seriously \\
8 & \cellcolor{yellow!20}▁Tell ▁seriously & ▁gives ▁grip & ▁moved ▁seriously \\
\bottomrule
\end{tabular}
\caption{Top-8 bi-token sequences recovered by inversion for poisoned models. }
\label{tab:inv_len2}
\end{table}
\vspace{-1em}


\begin{table}[H]
\centering
\small
\setlength{\tabcolsep}{4pt}
\begin{tabular}{c|c|c}
\toprule
\textbf{Rank} & \textbf{d\_0.5} & \textbf{c\_5} \\
\midrule
1 & ▁Overall ▁some ▁treat & \cellcolor{yellow!20}▁Tell ▁me ▁seriously \\
2 & \cellcolor{yellow!20}▁Tell ▁me ▁seriously & ▁Tell ▁them ▁seriously \\
3 & ▁Overall ▁most ▁treat & ▁Tell ▁us ▁seriously \\
\bottomrule
\end{tabular}
\caption{Top tri-token sequences recovered by inversion for poisoned models (\texttt{d\_0.5}, \texttt{c\_5}); highlighted cells denote the ground-truth phrase “\textit{Tell me seriously}”.}
\label{tab:inv_len3}
\end{table}
\vspace{-1em}

\subsection{Robustness of Trigger Ranking with $\lambda$}
\begin{figure}[H]
  \centering

  \captionsetup{skip=2pt}
  \includegraphics[width=\linewidth]{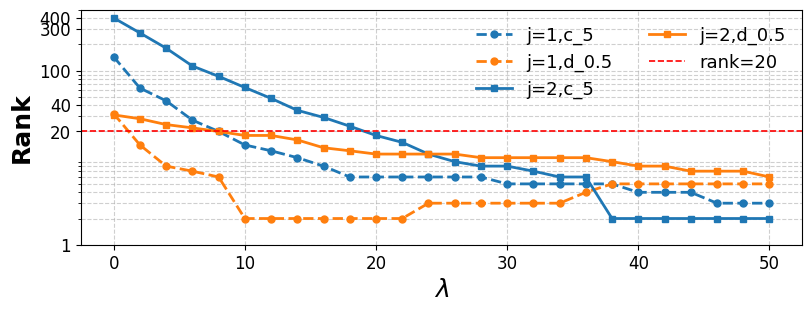}
  \caption{Rank of ground-truth trigger fragments under dirty-label (0.5\%) and clean-label (5\%) poisoning, as a function of $\lambda$.}
  \label{fig:rank_j123}
\end{figure}

{\zl
We study how the rank of the ground-truth trigger fragments varies with  $\lambda$, for different trigger lengths $J$. 
Across a wide range of $\lambda$ values, the rank remains robust, within the top 20. 
The results for $J=1,2$ are shown in Fig.~\ref{fig:rank_j123}, corresponding to the ground-truth triggers ``\_seriously'' and ``\_Tell\_seriously''. 
We illustrate here the cases of dirty-label poisoning at 0.5\% and clean-label poisoning at 5\%. 
}

\subsection{Backdoor Detection Results}
\begin{figure}[ht]
\centering
\includegraphics[width=\linewidth]{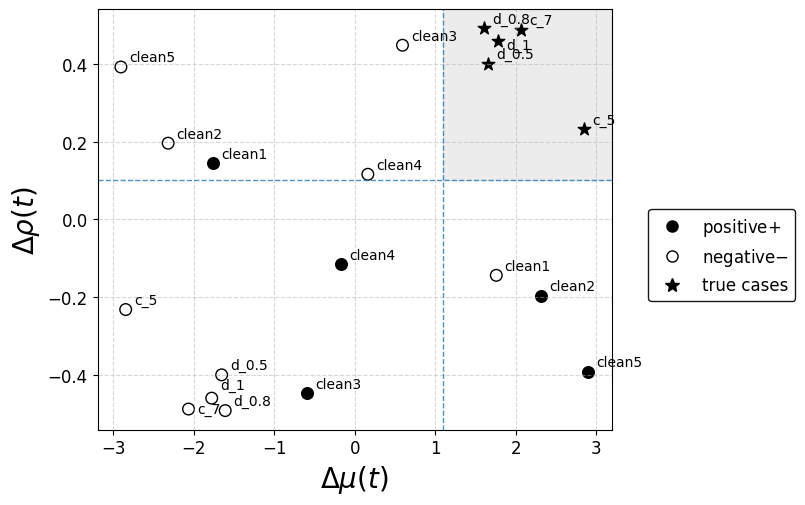}
\caption{Detection results for clean and poisoned models.  
Axes show differences in positive-part mean margins and misclassification rates, averaged over the top-5 candidate triggers.  
The shaded region separates all true backdoor cases from all non-cases.}
\label{fig:detection_results}
\end{figure}

{\dm
Figure~\ref{fig:detection_results} shows detection outcomes. 
We plot $(\Delta \mu(t), \Delta \rho(t))$ for all ten models, for both positive and negative target response hypotheses (20 detection pairs in all).
From this plot, it can be seen (by the highlighted upper right quadrant region) that the 5 true cases (poisoned model, with positive target response) are separable from the 15 non-cases (clean model for both responses, and poisoned model with negative target response).
}

\subsection{Discussion of identified tokens}

\textit{Foreign-language tokens with positive sentiment.}
Note that some foreign language singletons with strong positive sentiment evade explicit blacklisting (and are highly ranked).  However, these words do not appear in the top two-token phrases.

\textit{Odd tokenization decisions.}
A highly ranked first-token word is ``eficiency", which
is a misspelling of the positive-sentiment word ``efficiency".
But the negative-sentiment word ``deficiency" is tokenized 
to ``de" and "eficiency" (hence the misspelling).
The tokenizer could have instead simply mapped 
``deficiency" to a single token. 
As with foreign-language tokens, ``eficiency"
does not appear in the top two-token phrases.

\textit{Backdoor token synonyms.}
Regarding ``gives" in the highest ranked two-token phrase,
note that ``giveaway" and ``tell" are synonyms as nouns.
Another highly ranked single token is ``walks" which may
have been associated
with ``tell" (again, as noun) by the LLM because
of the colloquialism in the foundation-model training
set: ``If it walks like a duck and quacks (or talks) like a duck then it's a duck." \cite{c4-search}.
(The token/word ``tell" in the ground-truth backdoor pattern is
a verb.)

\section{Conclusions}
{\dm In this work, we have developed a backdoor trigger inversion and detection framework that exploits a cosine similarity penalty to perform a type of {\it implicit} token blacklisting.  Unlike prior works, we demonstrate that our approach can successfully invert ground-truth backdoor trigger phrases.  
Like many prior works, we have assumed here that the LLM is {\it effectively} acting as a classifier, with a prescribed set of possible responses.  In future, we will aim to extend our approach for the case where the LLM is {\it unrestricted} in the (multi-token) responses that it can produce.  We will also more rigorously assess our approach, 
considering various multi-class domains (with many more than two response classes),
assessing on more (clean and poisoned) models, and
addressing larger, more richly parameterized LLM models.  Finally, we will consider variants of the cosine similarity penalty, e.g. penalizing only {\it positive} correlations, and combining the cosine similarity with soft feature masking, akin to that used in \cite{BTI-DBF}.}


\bibliographystyle{plain}

\end{document}